# Enhancing End-to-End Multi-Task Dialogue Systems: A Study on Intrinsic Motivation Reinforcement Learning Algorithms for Improved Training and Adaptability

1st Navin Kamuni
*AI ML M.Tech*
*BITS Pilani WILP*
USA
navin.kamuni@gmail.com

2nd Hardik Shah
*Department of IT*
*Rochester Institute of Technology*
USA
hds6825@rit.edu

3rd Sathishkumar Chintala
*Department of IT*
*Fisher Investments*
USA
sathishkumarchintala9@gmail.com

4th Naveen Kunchakuri
*Department of IT*
*Eudoxia Resarch University*
USA
Knav18@gmail.com

5th Sujatha Alla
Old Dominion University,
Norfolk, Virginia
USA
salla001@odu.edu

*Abstract*— End-to-end multi-task dialogue systems are built with distinct modules for each step of the dialogue pipeline; one important module for handling user input is the policy module. The goal of this work is to improve these systems' action quality evaluation and training by exploring intrinsic motivation reinforcement learning algorithms. By concentrating on state visit frequency and promoting exploration through semantic similarity between utterances, it adapts random network distillation and curiosity-driven reinforcement learning. Findings on the MultiWOZ dataset show that systems based on intrinsic motivation perform better than those that depend on extrinsic rewards. In particular, remarkable average success rates of 73% were attained by systems using random network distillation trained on semantic similarity between dialogues, which is significantly higher than the baseline's average success rate of 60%. Optimizing Policy Proximal (PPO). In addition, performance indicators such as booking rates and completion rates show a 10% rise over the baseline. Furthermore, these intrinsic incentive models help improve the system's policy's resilience in an increasing amount of domains. This implies that they could be useful in scaling up to settings that cover a wider range of domains.

## I. INTRODUCTION

Human-computer interaction is a major area of focus for AI research, especially when it comes to creating conversational systems [1]–[8]. Reward scarcity in reinforcement learning poses a significant obstacle to the effectiveness and complexity of these systems, particularly task-oriented dialogue systems. In order to address these problems, this work introduces language input into training and investigates strategies for lowering reward sparsity. It explores random network distillation and curiosity-based reinforcement learning, providing dialogues with intrinsic rewards for novelty and encouraging policy exploration beyond simple models such as Proximal Policy Optimization.

## II. BACKGROUND: TASK-ORIENTED DIALOGUE SYSTEMS

Unlike conversational AI or chatbots, task-oriented dialogue systems are computer programs that employ natural language to assist users in achieving certain goals. Intent analysis, domain recognition, and slot filling are some of the phases these systems go through when users supply the information needed for things like taxi reservations. Until the user's wants are satisfied or the discussion is over, the process keeps going. Component interactions are coordinated by the conversation manager's conversation State Tracker (DST) and policy. To help express user goals, these systems make use of datasets that contain transcripts of human conversations divided into domains, intents, and slots.

The user simulator (US), dialogue manager, DST, policy, natural language generation (NLG), natural language understanding (NLU), and dialogue manager are important parts. The US is available in agenda-based and model-based forms, with the latter providing domain scalability and mirroring user goals [9], [10]. Using models such as RNNs, LSTMs, and BERT-NLU for intent categorization and slot value extraction, NLU focuses on transforming user utterances into conversation acts [11], [12]. With innovations like TripPy and TRADE, DST manages multi-domain complexity. Algorithms are used by policy components, which are essential for system-user interactions, in a Markov Decision Process framework [13]. NLG generates sentences directly from data, ranging from template-based to corpus-based approaches [17], [18]. Notably, issues with sparse reward and learning efficiency are addressed by pre-training protocols and various reward functions in task-oriented dialogue systems, like Q-learning and least-squares optimization [14] to [16]. Metrics for evaluation, such as information and success rates, evaluate the capabilities of the system; however, difficulties in conducting a thorough assessment still exist, particularly when poor US can mask policy performance [19]. The present study highlights the intricate components, procedures, and assessment challenges associated with task-oriented dialogue systems. It specifically focuses on user interaction simulation, natural language understanding, dialogue management, and the generation of human-like responses.

### III. RELATED WORKS

A number of novel approaches to improving discussion policies have been introduced in the field of dialogue systems through innovative studies. While study [21] created PETAL, a transfer reinforcement learning framework for personalizing dialogue systems by leveraging individual interactions and collective user data, study [20] introduced ACER, which uses reinforcement learning for efficient training of dialogue systems. Additionally, [22] introduced an algorithm that enhances the exploration efficiency of deep Q-learning agents in dialogue systems. [23] carried out a thorough evaluation of deep learning-based dialogue systems, covering model kinds, approaches, and future expectations. In order to improve learning from user interactions, study [24] presented a hybrid learning approach that combines imitation and reinforcement learning. Furthermore, [25] developed a dialogue management architecture based on feudal reinforcement learning. This architecture addresses scalability concerns by segregating decision-making into two categories: master policy selection and primitive action choices.

These methods do have some drawbacks, though. Long-term dependencies and complex dialogue contexts may be difficult for ACER [20]. PETAL [21] makes the assumption that individual interactions are good target domains for personalization, which may mean that it ignores a range of user preferences and the evolution of dialogue context. The exploration efficiency algorithm proposed by [22] needs to be validated in a variety of user interactions and confronts scaling issues. Generalization and adaptation problems limit the potential of deep learning-based dialogue systems [23]. Sensitive to the quality of the training data, the hybrid learning approach [24] needs to strike a balance between imitation and reinforcement learning. Finally, due to presumptions regarding domain ontology and information sharing, the Feudal reinforcement learning-based architecture [25] may face difficulties in real-world deployment. For dialogue systems to be robust and be used in real-world scenarios, these issues must be resolved.

### IV. EMPIRICAL STUDY

#### A. MultiWOZ Dataset

About 8000 human-to-human dialogue examples from an information center, spanning seven categories such as attraction, hotel, and restaurant, are included in the MultiWOZ dataset [26]. Task-oriented dialogue system research often makes use of it. The conversations, which range in type from single to multi-domain, typically center around two domains; over 30% of the conversations include a single domain, while three domains are the least prevalent. Annotations for starting aims, user utterances, and system conversation acts are included in the dataset.

#### B. Convlab-2

An improvement to Convlab [27], Convlab-2 is an open-source dialogue systems library created by [12] that facilitates the quick construction, training, and assessment of dialogue systems. It supports four main datasets: DealOrNoDeal [28], MultiWOZ [26], CamRest676 [17], and CrossWOZ [29]. Convlab-2 provides contemporary models for dialogue system components such as Natural Language Understanding (NLU), Natural Language Generation (NLG), and Dialogue State Tracking (DST). It supports both monolithic and pipeline dialogue systems. To reduce problems from upstream and downstream processes, the NLU module is not included in this study. Instead, utterance creation is done using a basic template NLG module. Additionally embedded is a simple rule-based DST module. ConvLab-2 is made to make it simple to set up and evaluate different approaches to dialogue systems, such as reinforcement learning methods. It has an environment module that links the system for reinforcement learning policy training and the user simulator (US), and it is built on OpenAI Gym [30]:

1. To sample an initial belief state, which includes the randomly chosen first user goal, the function *env.reset()* is utilized.
2. The function next state, reward, *done = env.step(action)* is used to take actions, retrieve instant rewards, advance the MDP to the next state, and determine whether the discussion has ended.

Moreover, independent of the environment variable used, the evaluator module can be activated within the environment to evaluate the efficacy of the interaction. When the dialogue fails *(done = False)*, the reward function is intended to offer a -1 reward; conversely, when the dialogue succeeds *(done = True and evaluator.task success = True)*, the reward function is intended to assign a positive reward of *L*. In ConvLab-2, the maximum number of turns in a conversation instance is denoted by *L* and is set to 40. Each phase's actions, which are often the Policy module's output, must be sent to the environment. The following policies are included in ConvLab-2:

1. A rule-based policy in which each user dialogue act is met with a tailored system reaction.
2. A Maximum Likelihood Estimation (MLE) policy that uses imitation to directly learn state-action pairs from the dataset.
3. PPO [31], Guided Dialog Policy Learning (GDPL) [12], and REINFORCE [32] are the three reinforcement learning policies.

The following performance metrics can be calculated using ConvLab-2's analyzer module:

1. Complete rate: The ratio of finished conversations to total conversations shows that, regardless of correctness, the system gave the user all the information they asked for:

$$Complete\ rate = \frac{completed\ dialogues}{total\ dialogues} \quad (1)$$

2. Success rate: The ratio of successful dialogues to all dialogues, where success denotes that all user requests were appropriately handled by the system:

$$Success\ rate = \frac{successful\ dialogues}{total\ dialogues} \quad (2)$$

3. Book rate: The ratio of booked dialogues that are successful to all of the dialogues that are bookable:

$$Book\ rate = \frac{successful\ bookings}{total\ number\ of\ bookable\ dialogues} \quad (3)$$

### C. Algorithms

PPO algorithm in the context of reinforcement learning, drawing parallels with Trust Region Policy Optimization (TRPO). PPO aims to control policy parameter updates to prevent drastic changes between iterations, akin to TRPO's approach of limiting the Kullback-Leibler (KL) divergence between the updated and old policies. However, PPO introduces a clipped surrogate function to address instability issues associated with TRPO. The surrogate function constrains the ratio of probabilities for actions under the updated and old policies, ensuring it remains within a predefined interval.

In exploration strategies, traditional methods like $\varepsilon$-greedy and entropy bonuses with more sophisticated approaches. Pseudo-count models, introduced by [33], estimate state visitation frequencies using density models, extending to high-dimensional spaces with hash mapping. Intrinsic motivation methods [34], [35] introduce novelty as a measure, rewarding less frequently visited states. In [36], intrinsic motivation is quantified as information gain via KL divergence. However, traditional reinforcement learning algorithms face challenges due to reward sparsity, prompting the introduction of intrinsic rewards. Episodic memory [37] appends large intrinsic reward states to facilitate learning from novel states. Curiosity-driven learning [38] employs an Intrinsic Curiosity (IC) module, comprising forward and inverse estimation models, to generate step-wise intrinsic rewards. Random Network Distillation (RND) [39] measures prediction error for novel states, encouraging exploration. In order to predict state features and actions, IC combines forward and inverse dynamics models, using feature vector differences as intrinsic rewards. It seeks to balance extrinsic and intrinsic rewards in order to maximize policy actions. RND promotes exploration by rewarding prediction errors and employing random neural networks to predict states. As their pseudocode and Algorithm 1 demonstrate, in reinforcement scenarios, both IC and RND can integrate with or separate from policy learning.

**Algorithm 1**
START
*Randomly initialize target and predictor network parameters, $\theta'_0$ and $\theta_0$;*
*Freeze $\theta'_0$;*
*Get initial state $s_0$;*
*Set $\eta$ and $\alpha$;*
*for i = 0, 1, 2,.....do*
　*Sample action $a_i$ given $s_i$;*
　*Compute embeddings $f_{\theta_i}(s)$ and $f_{\theta'_i}(s_i)$*
　*Optimise the loss $||f'_{\theta_i}(s_i) - f\theta_i(s)||^2$ w.r.t $\theta$ by gradient descent;*
　*Compute intrinsic reward $r_i^{int} = \eta||f'_{\theta_i}(s_i) - f\theta_i(s)||^2$;*
　$\eta \leftarrow (1-\alpha)\eta$
*for end*
END

### V. MODEL IMPLEMENTATION ENVIRONMENT

ConvLab-2 is an open-source library that has been updated since its February 2020 launch and is used in this study. It involves changes to a number of different parts, including models, NLG, and user agents. Models are trained and assessed in a manner consistent with reinforcement learning techniques, meaning that test and evaluation data are treated equally. ConvLab-2's DST uses a belief state encoder to process user goals and conversation history into fixed-length arrays. This covers the agent's unprocessed actions as well. Table II lists common hyperparameters for all models, while Table I shows data overviews and output array sizes. The batch size and discount factors were not changed during the hyperparameter tuning process; they remain at standard values. Due to resource and time constraints—which are explained in more detail in the following section—each experiment was restricted to one million training steps.

**TABLE I**
**THE INPUT AND OUTPUT OF THE BELIEF STATE AND ACTION ENCODERS**

| Input (structure) | Output (structure, dimension) |
|---|---|
| Belief state (dict) | (array, 340) |
| Action (dict) | (array, 209) |

**TABLE II**
**HYPERPARAMETERS**

| Hyperparameter | Value |
|---|---|
| Number of training steps | 1M |
| Discount factor | 0.99 |
| Dialogue sample batch size | 32 |
| Optimizer batch size | 32 |
| Optimizer model | AdamW ([31]) |
| Activation function | ReLU |

### A. PPO

The PPO algorithm is implemented similarly to Convlab-2, except for the hyperparameters. Following an analysis of the learning performance at different learning rates, optimizer models, and algorithm-specific hyperparameters, these modifications were made. In this implementation, the actor-critic architecture consists of two neural networks; hence, since there are no shared parameters, two different optimizers are used. Tables III and IV list the hyperparameters for the actor and critic models of PPO. The gradient for the actor loss is trimmed at a value of the same policy and is trained for five iterations of every sampled batch of conversations. All exploration comes from the models of intrinsic motivation because the entropy benefit is ignored. These PPO hyperparameters don't change while the RND and IC models are being trained.

**TABLE III**

**HYPERPARAMETERS FOR PPO (ACTOR)**

| Hyperparameter | Value |
|---|---|
| Actor input dimension | 340 |
| Actor hidden dimension | 100 |
| Actor output dimension | 209 |
| Learning rate | 5e-6 |
| Surrogate clip ($\eta$) | 0.1 |
| GAE ($\lambda$) | 0.95 |

**TABLE IV**
**HYPERPARAMETERS FOR PPO (CRITIC)**

| Hyperparameter | Value |
|---|---|
| Critic input dimension | 340 |
| Critic hidden dimension | 50 |
| Critic output dimension | 1 |
| Learning rate | 1e-5 |

### B. Random Network Distillation

Algorithm 1 provides two operational modes for the functionality of the RND model, guiding its implementation. The model can calculate step-wise intrinsic rewards using NLU (called Utt, for example, RND (Utt)) or dialogue acts (called DAs, such RND (DAs)) that are exchanged between the user and the system. The RND model's two modes go through warm-up stages before working together to train the policy.

#### 1) Random Network Distillation from Dialogue Acts

When using dialogue acts to produce intrinsic rewards, two feed-forward neural networks are used: one for the target model, which has fixed parameters throughout training, and another for the predictor model. Table V contains information about the hyperparameters for these networks.

**TABLE V**
**HYPERPARAMETERS FOR TARGET NETWORKS AND RND (DAS) PREDICTION**

| Hyperparameter | Value |
|---|---|
| Input dimension | 418 |
| Hidden dimension | 524 |
| Output dimension | 340 |

#### 2) Random Network Distillation from Utterances

When using utterances for intrinsic reward creation, the predictor and target models are trained using a pre-trained BERT model. Although the entire pre-trained BERT model can be optimized during training, the effect of freezing all BERT layers—aside from the last layer, the classification head, which is learned simultaneously with the policy—was studied. Even with comparable outcomes, the earlier approach requires far more training wall time. As a result, it was decided to train the BERT classifier head in all scenarios. Table VI provides a full breakdown of the hyperparameters for the BERT-based predictor and target models.

**TABLE VI**
**HYPERPARAMETERS FOR RND ($U_{TT}$) PREDICTOR AND TARGET NETWORKS**

| Hyperparameter | Value |
|---|---|
| Critic input dimension | BertForSequenceClassification |
| Critic hidden dimension | bert-base-uncased |
| Critic output dimension | 512 |
| Learning rate | 340 |

#### 3) Random Network Distillation General Hyperparameter

Table VII lists the hyperparameters used for RND in this work, which includes both modes that use DAs and utterances. There are notable differences in some hyperparameters between RND (DAs) and RND (Utt), all of which have been painstakingly adjusted via a grid search procedure.

**TABLE VII**
**HYPERPARAMETERS FOR RND (DAS AND UTT) TRAINING**

| Hyperparameter | Value |
|---|---|
| Learning rate (both) | 1e-3 |
| Warmup episodes (DAs) | 100 |
| Warmup episodes (Utt) | 200 |
| Moving average steps period (DAs) | 2 |
| Moving average steps period (Utt) | 10 |
| Update rounds (DAs) | 5 |
| Update rounds (Utt) | 1 |
| RND gradient clip value (DAs) | 10 |
| $r^{int}$ multiplier initial value (Utt) | 1.0 |
| $r^{int}$ multiplier initial value (both) | 5.0 |
| $r^{int}$ multiplier initial value (both) | 0.001 |
| $r^{int}$ linear annealing steps (DAs) | 20000 |
| $r^{int}$ linear annealing steps (Utt) | 50000 |

### C. Intrinsic Curiosity

There are two modes of operation for intrinsic curiosity (IC): IC (DAs) and IC (Utt). Whereas IC (Utt) processes the tuple (state, action, future state) as dialogue acts, IC (DAs) employs NLG to generate utterances from conversation acts. As outlined in Algorithm 1, IC provides two training methods: combined optimization with the policy and pre-training with a predefined policy for intrinsic incentives. The specific hyperparameters needed for each of these approaches are listed in the Results section. The IC models are feed-forward neural networks with ReLU activations; IC (Utt) uses a BERT base uncased tokenizer, and IC (DAs) uses a two-layer network to encode dialogue actions. In pre-training, the intrinsic reward is determined by scaling the forward model's loss with hyperparameter $\eta$.

**TABLE VIII**
**HYPERPARAMETERS FOR IC (DAS AND UTT) MODEL**

| Hyperparameter | Value |
|---|---|
| Number of steps for pre-train | 1000 |
| Learning rate (pre-train) | 1e-3 |
| Learning rate (joint) | 1e-5 |
| Update rounds | 1 |
| Gradient clipping parameter | 10 |
| $\eta$ | 0.01 |
| $\beta_{DAs}$ | 0.2 |
| $\beta_{Utt}$ | 0.2 |
| $\beta_{joint}$ | 0.8 |
| $\lambda_{pol}$ | 0.5 |
| Inverse model hidden dimension | 524 |
| Forward model hidden dimension | 524 |
| State encoder output dimension | 256 |
| State encoder max length | 200 |

## VI. RESULT ANALYSIS

Since Convlab-2's training does not provide direct performance monitoring, an Analyzer wrapper is used to evaluate dialogue sample size in real-time using a hyperparameter. Although it takes longer overall, simultaneous training and evaluation are essential for hyperparameter tuning. Less dialogue samples increase training speed but decrease accuracy. 1000 random dialogues are used in this study to assess performance in a consistent and trustworthy manner.

### A. Baseline Models

Within its system policy module, Convlab-2 provides four reinforcement learning algorithms; Table IX shows the average success rates. Although GDPL [40] is regarded as the best model in terms of performance on the MultiWOZ dataset in [12], the authors of Convlab-2 report that PPO has a success rate of 0.74 and GDPL has a measly 0.58, with the latter just outperforming MLE and REINFORCE. Additionally, using the same dataset, the success rate stated in [40] is 0.59. This suggests that there may be problems with specific library components or that the default training parameters recommended by Convlab-2 aren't the best ones.

**TABLE IX**
**CONVLAB-2 BASELINE MODEL SUCCESS RATE**

| Model | Success rate |
|---|---|
| MILE | 0.56 |
| REINFORCE | 0.54 |
| PPO | 0.74 |
| GDPL | 0.58 |

### B. Effect of the Number of Sampled Dialogues in Evaluation

Here, we provide empirical evidence in support of the qualitative insight that any taught policy can be evaluated by fixing the number of dialogue samples, or $n_{eval}$. The success rate and standard deviation during the evaluation of a PPO model that was trained for 1 million steps while modifying $n_{eval}$ are shown in Fig. 1. For every $n_{eval}$, twenty calculations of the success rate were made. The results show that the performance measures vary very little when $n_{eval} = 1000$ is set. As a result, we determine $n_{eval} = 1000$ for all experiments to provide consistent and extremely trustworthy evaluations of model performance during training.

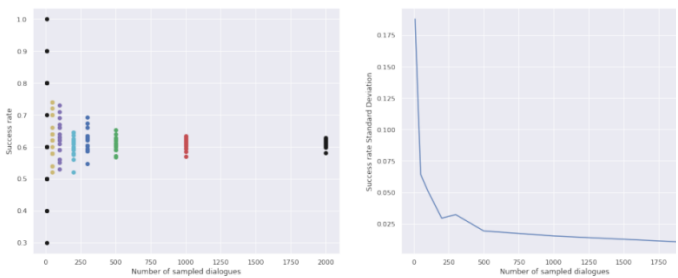

Fig. 1. Effect of the number of sampled dialogues in evaluation

## VII. CONCLUSION & FUTURE WORKS

Using the MultiWOZ dataset and Convlab-2, this study examined intrinsic incentive models with an emphasis on RND and IC strategies to improve dialogue system performance. The system policy performance was significantly enhanced by RND, especially with its utterance-based approach, which increased the success rate from 60% to 73% and increased the completion and book rates by 10%. However, the study's dependence on Convlab-2 and the ease of use of the dialogue system components presented challenges. Future developments will involve upgrading the user simulator, implementing more complex NLG and DST algorithms, and resolving problems with Convlab-2, such as user objective assignment and performance metric computations. Future research aims to improve user interfaces for a wider audience, customize interactions to users' accessibility needs, and use intrinsic motivation reinforcement learning to multi-task dialogue systems for web accessibility [41, 42].

## VIII. DECLARATIONS

*A. Funding:* No funds, grants, or other support was received.

*B. Conflict of Interest:* The authors declare that they have no known competing for financial interests or personal relationships that could have appeared to influence the work reported in this paper.

*C. Data Availability:* Data will be made on reasonable request.

*D. Code Availability:* Code will be made on reasonable request.